\documentclass{article}

\usepackage{microtype}
\usepackage{graphicx}
\usepackage{subcaption}
\usepackage{booktabs} % for professional tables
\usepackage{float}

\usepackage{hyperref}

\usepackage[accepted]{icml2021}

\icmltitlerunning{Decadal Forecasts with ResDMD: a Residual DMD Neural Network}

\begin{document}

\twocolumn[
\icmltitle{Decadal Forecasts with ResDMD: a Residual DMD Neural Network}



\icmlsetsymbol{equal}{*}

\begin{icmlauthorlist}
\icmlauthor{Eduardo Rodrigues}{ibm}
\icmlauthor{Bianca Zadrozny}{ibm}
\icmlauthor{Campbell Watson}{ibm}
\icmlauthor{David Gold}{ibmgbs}
\end{icmlauthorlist}

\icmlaffiliation{ibm}{IBM Research}
\icmlaffiliation{ibmgbs}{IBM Global Business Services}

\icmlcorrespondingauthor{Eduardo Rodrigues}{edrodri@br.ibm.com}

\icmlkeywords{Dynamic Mode Decomposition, Resnet, Sea surface temperature, Decadal, Linear Inverse Model}

\vskip 0.3in
]



\printAffiliationsAndNotice{}  % leave blank if no need to mention equal contribution

\begin{abstract}
    Operational forecasting centers are investing in decadal (1-10 year) forecast systems to support long-term decision making for a more climate-resilient society. One method that has previously been employed is the Dynamic Mode Decomposition (DMD) algorithm -- also known as the Linear Inverse Model -- which fits linear dynamical models to data.
    %It uses the computationally efficient singular value decomposition (SVD) algorithms to
    %characterize the dynamics of a system in terms of spatio-temporal coherent structures.
   While the DMD usually approximates non-linear terms in the true dynamics as a linear system with random noise, we investigate an
    extension to the DMD that explicitly represents the non-linear terms as a neural network. Our weight initialization allows the network to produce sensible results before training and then improve 
    the prediction after training as data becomes available. In this short paper, we evaluate the proposed architecture for simulating global sea surface temperatures and compare the results with the standard DMD and seasonal forecasts produced by the state-of-the-art dynamical model, CFSv2.
\end{abstract}

\section{Introduction}

In recent years there has been significant effort by major dynamical modeling groups to perform decadal (1-10 year) forecasts (e.g., \citet{yeager2018predicting} and \citet{flor}). These forecasts 
can bridge the gap from seasonal forecasts \cite{kirtman2014north} to multi‐decadal predictions \cite{eyring2016overview} and are specifically designed to forecast fluctuations through knowledge of the current climate state and multi-year ocean variability. Although skill is limited by the predictability of phenomena like the El Ni{\~{n}}o–Southern Oscillation (ENSO), decadal forecasts can provide critical and actionable information on regional climate trends.

Another strategy for decadal predictions that has received attention 
uses data to fit a statistical model. A particular example of such an approach
is found in \cite{newman2013empirical}, in which the Linear Inverse Model, also known as Dynamic 
Mode Decomposition (DMD), is applied to make decadal forecasts. DMD has been used to successfully model climatic phenomena such as tropical diabatic heating and sea surface temperatures (SSTs; \citet{Huddart2017}). In its most basic form, DMD seeks to fit a linear dynamical system of the form:

\begin{equation}
\frac{d}{dt}\mathbf{x}=\mathbf{A x}
\end{equation}

\noindent
which has an exact solution given by:

\begin{equation}
\mathbf{x}(t_0+t)=\mathit{e}^{\mathbf{A}\mathit{t}}\mathbf{x}(t_0)
\end{equation}

\noindent
The dynamics are characterized by the eigenvalues and eigenvectors of $\mathbf{A}$, typically
the leading eigenvalue/eigenvector pairs.

A common strategy in modeling environmental variables with DMD is to assume that the non-linear
dynamics can be approximated by a linear system with random noise, usually Gaussian. In this paper, we seek a different strategy in which we explicitly model the non-linear term of the 
dynamics with a neural network. In particular, our neural network resembles a Resnet \cite{he2016deep}
with the difference that, instead of the identity, the bypass is the DMD solution. Initializing 
the non-linear path with small random weights makes the untrained network predict the DMD
solution. After training, however, results improve compared with this baseline (similar to \cite{8588749}).

In the next sections, we present our proposed approach, which we call the Residual DMD Neural Network (ResDMD), and apply it to predict sea surface temperature (SST), comparing it to the standard DMD and a state-of-art dynamical model, CFSv2. 

%\We believe that models such as this can improve decadal forecasts, and they can be combined with dynamical models to improve skill for long range forecasts.

%\section{Related work}

\section{Background and Contributions}

In this section, we describe the architecture of our network and the rationale for its structure.
We define what the initialization is and the benefit of this choice. Finally, we describe the
training procedure, which, in the current stage of this work, could be further optimized, but
we leave this as future work.

A multidimensional environmental variable $\mathbf{x}$ (typically an anomaly with respect to the long-term
average, i.e. the climate mean), whose components extend over the spatial domain, can be modeled as:

\begin{equation}
    \frac{d\mathbf{x}}{dt} = L\mathbf{x} + N(\mathbf{x})
    \label{eqA}
\end{equation}

\noindent
where $L$ is the linear part of the dynamics, and $N$ represents nonlinear terms.
$N$ is usually, however, approximated by a linear operator, as in \cite{winkler2001linear}.
This approach has the major benefit of easiness to compute, since it relies on the computationally efficient singular value decomposition (SVD) algorithm. Nonetheless, one can only
fully understand the system statistics by knowing the form of $N$.

Here, we seek to approximate the nonlinear term $N$ as a neural network trained with 
stochastic gradient descent (SGD). We leverage, however, the structure and procedure for
training the linear part from previous work (such as \cite{winkler2001linear} and
\cite{newman2013empirical}). The procedure for training the linear part is the Dynamic
Mode Decomposition (DMD), also known as Linear Inverse Model in the atmospheric sciences.

One of the greatest advantages of the DMD method is to project the dynamics of the system being studied
into a low-dimensional space, spanned by a set of spatially correlated
modes that have the same linear behavior in time \cite{kutz2016dynamic}. In discrete 
form the solution is given by:

\begin{equation}
    \mathbf{x}(t_0+n*\Delta t)=\Phi \Lambda^n \Phi^{\dagger}\mathbf{x}(t_0)
\end{equation}

\begin{figure}[htbp]
\centerline{\includegraphics[width=0.4\textwidth]{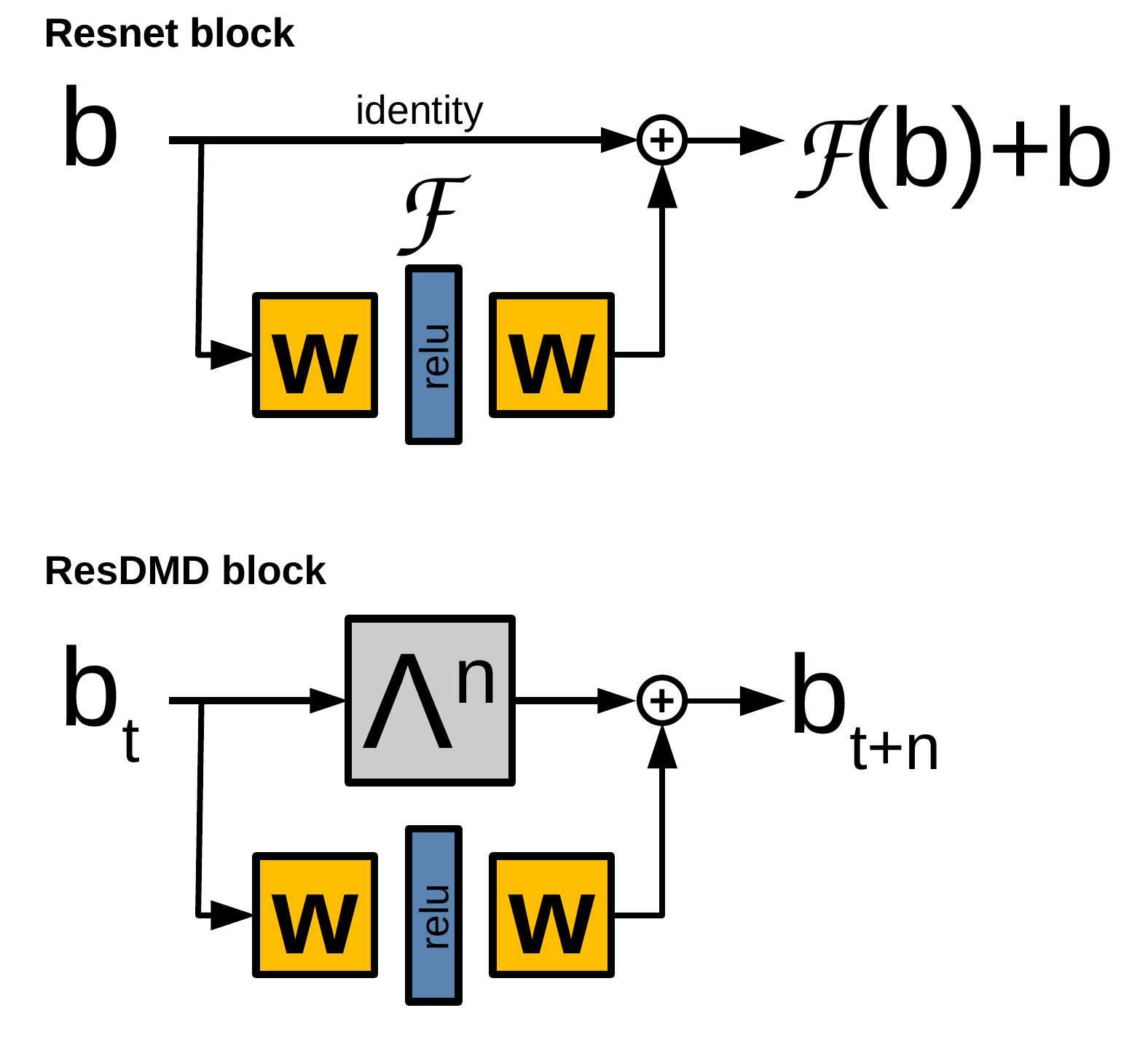}}
\caption{Comparison between Resnet block and ResDMD block.}
\label{fig1}
\end{figure}

\noindent
in which $\Phi$ and $\Lambda$ are the eigenvectors and eigenvalues of the matrix ($A$) approximating
the system in Equation \ref{eqA}. The pseudo inverse of $\Phi$, i.e. $\Phi^{\dagger}$, projects
the state variable $\mathbf{x}$ into a low dimension space. Its dimension size is given by the leading
eigenvalues of $A$, whose number is a hyperparameter of the method.

Our proposed model also operates initially in the same low dimensional space as the DMD model. Its
first step is to use the matrix $\Phi$ to project the initial state vector, which in our case extends over the whole globe:

\begin{equation}
    \mathbf{b} = \Phi^{\dagger}\mathbf{x}
\end{equation}

\noindent
This approach has the advantage of reducing the risk of overfitting with high dimensional data as
the dynamics are restricted to this low dimensional sub-space. The projection matrix is 
an optimizable parameter in our model which can be further optimized to best fit observations 
during the SGD procedure.

In order to approximate the non-linear terms of the dynamics, we use a neural network architecture (with parameters $\mathbf{w})$ 
similar to the Resnet \cite{he2016deep}. The key assumption of Resnet is that it is easier to optimize the
residual mapping than to optimize the original unreferenced signal. The reference in that neural network 
is the signal itself, i.e. an identity mapping (Figure \ref{fig1} top). Our ResDMD architecture is similar, but
the reference in our case is the DMD solution (Figure \ref{fig1} bottom).

Initializing the weights of the projection matrices and $\Lambda$ with the DMD matrices, and initializing
the weights $\mathbf{w}$ to small random values effectively makes the SGD procedure look for a solution with a non-linear additive term close to the DMD solution. As one accumulates more and more observation data and
trains the model, the network better approximates equation \ref{eqA}. In the next section, we evaluate
ResDMD with SST forecasts compared to the standard DMD procedure and the CFSv2 model.

Our model can be further stacked in larger networks, similarly to the very deep Resnet models. Additional
restrictions, such as shared parameters between layers for instance, can be added. In this paper, we only show results with a single block to test the viability of the method. 

The training procedure is basically divided in two stages. First, we fit a DMD model to the observed
data (training set). The resulting matrices $\Phi$ and $\Lambda$ are then used to initialize the corresponding
weights of the neural network. In addition, the weights $w$ are initialized with small random values so that
the untrained network produces the DMD solution. The second stage corresponds to the SGD optimization procedure, in which we run through the training set again. In this stage, all parameters ($\Phi$, $\Lambda$ and $w$) are updated. 

\section{Experiments}

In order to test the performance of the ResDMD, we trained and tested it with sea surface temperature (SST). This
variable is known to be an important predictor for regional climate trends and consequential weather events, and it is reasonably
slow varying making it a good candidate for a first target variable. We hereby compare
performance of the ResDMD with the standard DMD and CFSv2 seasonal forecast ensemble. 

The SST dataset used in our experiments is the Extended Reconstructed Sea Surface Temperature, Version 5 (ERSSTv5)
\cite{huang2017extended}. It provides estimates of SSTs from 1854 to present and is most reliable
from 1950. Its spatial resolution is two degrees and is available monthly. We also use the CFSv2 seasonal forecasts as a benchmark for our predictions; it was not used as input to our model.
%We believe however that the best
%approach for a more reliable predictor is to combine data-driven, as our model, and dynamical models. In this
%work, however, we do not explore this approach.

Instead of directly predicting absolute values, a common practice in long range forecasts of SST and other climate variables is to predict anomalies from the expected value. (The strong seasonality in this data can bias the evaluation metric.) We computed anomalies by subtracting a monthly climate mean (from the period 1980 to 2010) from the corresponding training and testing sets. For ERSSTv5, the training set ranged from 1850-2010 and the test set from 2011 to 2020. In addition, we computed
CFSv2 forecast anomalies using the CFSv2 forecast climatologies from 1980 to 2010. This extra 
precaution aims to make the comparison with CFSv2 more realistic -- computing forecast anomalies from 
observed climate means (in this case ERSSTv5) could hurt the CFSv2 performance by exposing model biases \cite{kumar2012analysis} and confer an unfair advantage to our model.

After training the standard DMD model and ResDMD with the training set, we evaluated the performance in 
the test set using the anomaly correlation coefficient (ACC), which is commonly used to measure the skill
of a forecast system. 
%Its definition is:
%
%\begin{equation}
%    ACC = \frac{\overline{(f-c_f)(o-c_o)}}{\sqrt{\overline{(f-c_f)^2}\overl%ine{(o-c_o)^2}}}
%\end{equation}
%
%\noindent
%Where $f$ is the forecast, $c_f$ is the forecast climatlogy, $o$ stands for observations, and finally $c_o$
%is the climate mean.
ACC values closer to unity indicate higher skill of the forecast system. We also
used an ACC difference ($\Delta$ACC) to compare both the standard DMD and ResDMD to CFSv2. Positive 
values of $\Delta$ACC indicate the corresponding DMD model is better than CFSv2, and negative values indicate
the opposite.

\section{Results}

\begin{figure}[t!]
\centering
\begin{subfigure}[b]{0.47\textwidth}
   \includegraphics[width=1\linewidth]{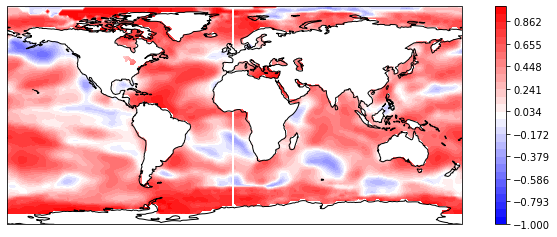}
   \caption{Standard DMD}
   \label{fig:1111} 
\end{subfigure}
\begin{subfigure}[b]{0.47\textwidth}
   \includegraphics[width=1\linewidth]{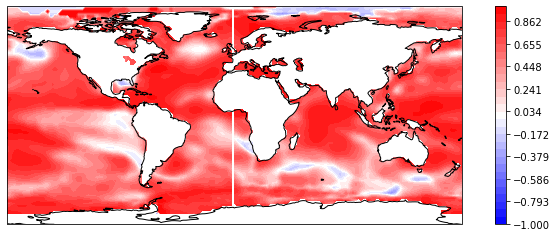}
   \caption{ResDMD}
   \label{fig:1112}
\end{subfigure}
\caption{ACC of SST forecast of 6 months lead time over the 2010 to 2020 period}
\label{fig:fig111}
\end{figure}

\begin{figure}[b!]
\centering
\begin{subfigure}[b]{0.47\textwidth}
   \includegraphics[width=1\linewidth]{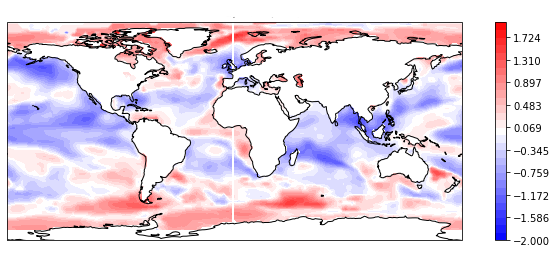}
   \caption{Standard DMD}
   \label{fig:2221} 
\end{subfigure}
\hfill
\begin{subfigure}[b]{0.47\textwidth}
   \includegraphics[width=1\linewidth]{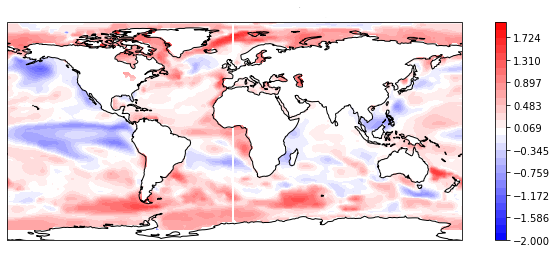}
   \caption{ResDMD}
   \label{fig:2222}
\end{subfigure}
\caption{$\Delta$ACC of SST forecast of 9 months lead time over the 2011 to 2020 period compared to CFSv2}
\label{fig:fig222}
\end{figure}

In this section, we present some of the ACC maps we obtained in order to compare ResDMD with the standard DMD and CFSv2. The number of points for each spatial coordinate in those maps corresponds to the number of years in our test set (10 years) times twelve months.

Figure \ref{fig:fig111} shows ACC values for the standard DMD (top) and ResDMD (bottom) models. This particular example corresponds to predictions 6 months in advance (6-month lead time). Overall, ResDMD shows higher skill than the standard DMD. In particular regions of low correlation, such as the southern Atlantic Ocean, southern Indian Ocean and eastern Pacific Ocean, ResDMD improves skill considerably compared to the standard model. However, skill in the eastern tropical Pacific (where ENSO dynamics are observed) does not improve as much. In order to better understand this behaviour we compare results to the CFSv2 ensemble forecasts.

Figure \ref{fig:2221}--\ref{fig:2222} presents the $\Delta$ACC for the standard DMD ACC and CFSv2, and ResDMD and CFSv2, respectively. Positive values indicate the DMD models are more skillful than the CFSv2. Here, we are only showing $\Delta$ACC for 9-month lead times. Again, overall one can see that ResDMD performs significantly better than the standard model. In addition, ResDMD also performs better in many places than the CFSv2. However, ResDMD has less skill in the eastern Pacific/ENSO region compared to CFSv2, a pattern that is also observed at other lead times (not shown). A possible explanation for the lack of skill in this region is that the dynamics of the state variable -- explicitly simulated by CFSv2 -- may be driven by a forcing term, which is strong enough to change the internal dynamics of this variable. In order to tackle this issue, we are exploring additional predictors such as 150-mb geopotential height.

\begin{figure}[h!]
\centering
\begin{subfigure}[b]{0.47\textwidth}
   \includegraphics[width=\textwidth]{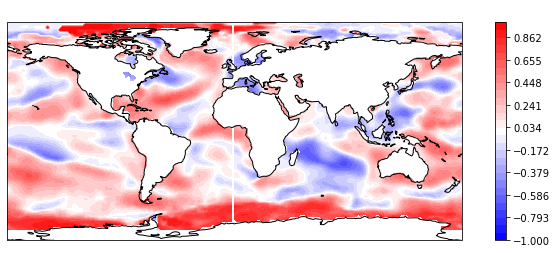}
   \vspace{-0.7cm}
   \caption{Standard DMD - 1 year lead time}
   \label{fig:3331} 
\end{subfigure}
\hfill
\begin{subfigure}[b]{0.47\textwidth}
   \includegraphics[width=\textwidth]{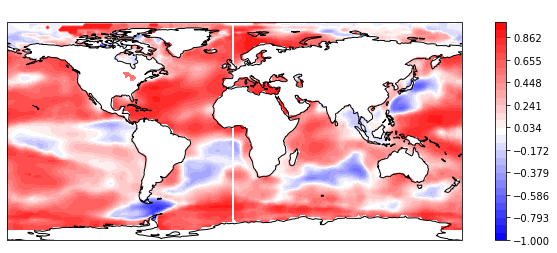}
   \vspace{-0.7cm}
   \caption{ResDMD - 1 year lead time}
   \label{fig:3332}
\end{subfigure}

\vspace{-0.17cm}
\begin{subfigure}[b]{0.47\textwidth}
   \includegraphics[width=\textwidth]{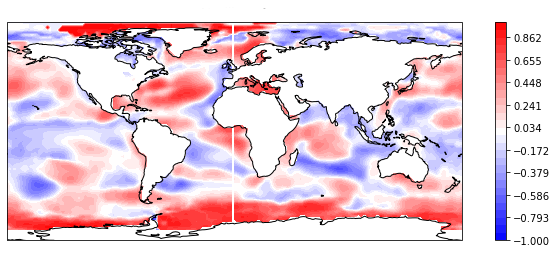}
   \vspace{-0.7cm}
   \caption{Standard DMD - 5 years lead time}
   \label{fig:3333}
\end{subfigure}
\hfill
\begin{subfigure}[b]{0.47\textwidth}
   \includegraphics[width=\textwidth]{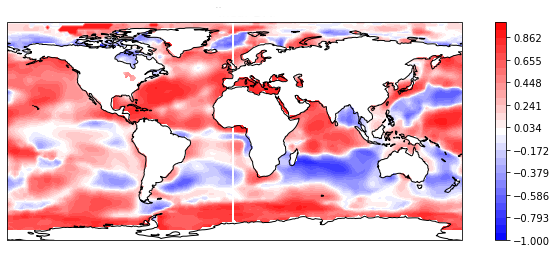}
   \vspace{-0.7cm}
   \caption{ResDMD - 5 years lead time}
   \label{fig:3334} 
\end{subfigure}

\vspace{-0.17cm}
\begin{subfigure}[b]{0.47\textwidth}
   \includegraphics[width=\textwidth]{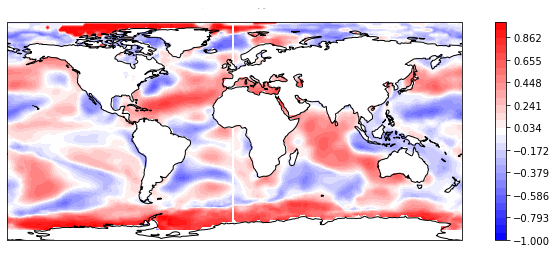}
   \vspace{-0.7cm}
   \caption{Standard DMD - 10 years lead time}
   \label{fig:3335}
\end{subfigure}
\hfill
\begin{subfigure}[b]{0.47\textwidth}
   \includegraphics[width=\textwidth]{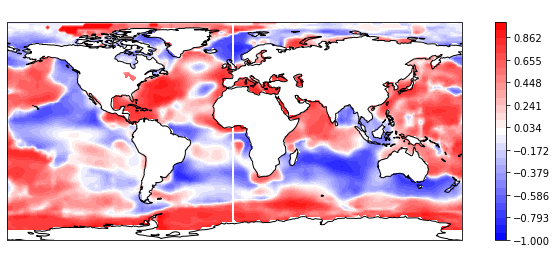}
   \vspace{-0.7cm}
   \caption{ResDMD - 10 years lead time}
   \label{fig:3336}
\end{subfigure}

\caption{ACC of SST forecast }
\label{fig:fig333}
\end{figure}

Finally, we present longer range forecasts by the standard DMD and ResDMD (Figure \ref{fig:fig333}) For 1-year lead time, the proposed ResDMD clearly performs better than the standard DMD across most of the globe. For 5-year lead time, the ResDMD error increases is specific locations such as the southern Indian Ocean, however it remains more skillful overall than the standard DMD. The ResDMD errors continue to amplify with lead time, and we conjecture that at longer lead times (e.g., 7-10 years) the ResDMD is no longer appropriate. We are actively exploring deeper neural networks with additional regularization for these longer lead times which will have the effect of damping the forecast close to the climate mean, however.

\section{Final remarks}
In this paper, we proposed a novel extension to the DMD method (ResDMD) that models non-linear additive terms of a dynamical system as a neural network.
%ResDMD is similar to the Resnet model, but the bypass is the DMD solution.
Weight initialization is such that, before the SGD training, the prediction is already a sensible forecast. We have shown predictions of global SSTs compared to the standard DMD and CFSv2, with notable improvements in some regions. 

The objective of the ResDMD model is to make decadal forecasts which are becoming increasingly important for climate change policy and to underpin other decision-making. We believe that data-driven methods, such the one proposed here, will be an important tool along with physically-driven methods to perform such forecasts. We expect to combine these two types of methods (i.e., a hybrid physical-statistical approach) in forthcoming work to address the shortcomings of the ResDMD approach.

\clearpage
\bibliographystyle{icml2021}
\bibliography{ref}

\end{document}